# Clubs-based Particle Swarm Optimization

Wesam Elshamy, Hassan M. Emara and A. Bahgat

Department of Electrical Power and Machines,

Faculty of Engineering, Cairo University, Egypt

{wesamelshamy, hmrashad, ahmed.bahgat}@ieee.org

**Abstract** – This paper introduces a new dynamic neighborhood network for particle swarm optimization. In the proposed Clubs-based Particle Swarm Optimization (C-PSO) algorithm, each particle initially joins a default number of what we call 'clubs'. Each particle is affected by its own experience and the experience of the best performing member of the clubs it is a member of. Clubs membership is dynamic, where the worst performing particles socialize more by joining more clubs to learn from other particles and the best performing particles are made to socialize less by leaving clubs to reduce their strong influence on other members. Particles return gradually to default membership level when they stop showing extreme performance. Inertia weights of swarm members are made random within a predefined range. This proposed dynamic neighborhood algorithm is compared with other two algorithms having static neighborhood topologies on a set of classic benchmark problems. The results showed superior performance for C-PSO regarding escaping local optima and convergence speed.

## I. INTRODUCTION

The relatively recent invention of computers with its increasing computational power allowed researchers to implement myriads of paradigms and algorithms and verify their efficiency. Researchers developed many algorithms inspired from natural phenomena such as the annealing in metallurgy [1], biological processes such as in genetics [2] and the immune system [3], animals and insects behaviors such as ants [4] and birds [5] and even cultures [6].

Particle Swarm Optimization (PSO) [5] is among these nature inspired algorithms. It is inspired by the ability of birds flocking to find food that they have no previous knowledge of its location. Every member of the swarm is affected by its own experience and its neighbors' experiences. Although the idea behind PSO is simple and can be implemented by two lines of programming code, the emergent behavior is complex and hard to completely understand [7].

In this paper we propose and study a new dynamic social network for PSO, where the social networks of the best performing particles shrink to reduce their influence on other particles, while the social networks of the worst performing particles expand to allow them to learn from more particles.

The organization of this paper is as follows; in part II we present the basic version of PSO and some of its variants and related work, followed by an explanation of the proposed Clubs-based PSO (C-PSO) algorithm in section III, while section IV presents the experiments to be conducted. The experiments' results are presented in section V, then a conclusion follows in section VI.

## II. PARTICLE SWARM OPTIMIZATION

Particle Swarm Optimization was inspired by the ability of a flock of birds or a school of fish to capitalize on their collective knowledge in finding food or avoiding predators. Each swarm member or particle has a small memory that enables it to remember the best position it found so far and its goodness. Particles are affected by their own experience (best found position) and their neighbors' experiences (best found position by the neighbors). The behavior of particles is described by (1) and (2).

$$v_{id}(t+1) = w \times v_{id}(t) + lrn_1 \times rand_1 \times (p_{id}(t) - x_{id}(t)) \quad (1)$$
$$+ lrn_2 \times rand_2 \times (p_{gd}(t) - x_{id}(t))$$

$$x_{id}(t+1) = x_{id}(t) + v_{id}(t+1) \quad (2)$$

In (1), $v_{id}$ is the speed of particle $i$ in dimension $d$. The first right hand side term corresponds to the inertia force that pushes the particle in its old direction, where $w$ is the weight value that controls this inertia force. The second term corresponds to the cognitive or personal experience component. It attracts the particle from its current position $x_{id}$ to its best found position so far in that dimension $p_{id}$ affected by a learning weight $lrn_1$ and a uniformly distributed random variable $rand_1$ in the range (0, 1). The third term corresponds to the social influence of the neighbors on the particle. It affects the particle by attracting it from its current position $x_{id}$ to the best position found by its neighbors $p_{gd}$ and this influence is controlled by a learning weight $lrn_2$ and another independent random variable $rand_2$ uniformly distributed in the range (0, 1). For each time step, as described by (2), each particle moves by a step of value $v_{id}$ in the $d^{th}$ dimension.

The PSO algorithm itself has evolved. The weight parameter $w$ was not included in the basic algorithm. It was added later and researchers examined the effect of varying its value [8]. A speed limit for the particles was introduced to prevent the explosion of speed values.

PSO operates in three spaces, the social network space, the parameter space of problem variables and the evaluative space [7] where estimates for the goodness of solutions are defined. Various social networks have been proposed and investigated by researchers [9]. In the original PSO algorithm, the social network connects every particle to all other particles and it is only influenced by the one that has the best experience compared to all particles. We will refer to this algorithm as PSO-g where 'g' stands for global. Though this algorithm converges rapidly, it could get easily trapped in local minima. After the particles are initialized, the first best found position by all particles attracts them all and as long as the particles experience improving performance in their new positions while heading for this first found minimum, they get more strongly attracted towards this local minimum.

A variant of the simple PSO has a ring social network. In this algorithm the particles are arranged in an imaginary ring and every particle is connected to its immediately preceding and succeeding particles in this ring. We will refer to this algorithm as PSO-l where 'l' stands for local. This algorithm converges slower than PSO-g but it is less susceptible to local minima and enjoys a higher degree of particles diversity. The influence of each particle in the swarm is limited to its two immediate neighbors. This influence limitation helps the particles to explore the search space with different points of attraction instead of a single best found point in the PSO-g algorithm. On the other hand, it may lead to excessive wandering for the particles leading to slow convergence even in easy problems having single optimum.

*Related Work:*

Both PSO-g and PSO-l are based on a static neighborhood network. Because the first stages of the search for the global best position require exploration of possible solutions, which PSO-l can do better. While later stages require exploitation of the best found candidate solutions by early stages of the search, which PSO-g is clever at. Researchers suggested using a dynamic neighborhood.

In [10], the neighborhood of each swarm member expands from an initial network that connects each particle to itself at early stages of the search, to a network that fully connects it to all other particles. This algorithm transforms gradually from acting like PSO-l in early stages of the search, to behave more like PSO-g at late stages. Two network expanding procedures have been introduced. Both of them depend on the current position of the particle to search for nearby particles to add to its neighborhood list.

In [11], a Fitness-Distance-Ratio based PSO (FDR-PSO) algorithm is introduced. In this algorithm, each particle is affected by three components; the cognitive, social and the FDR components. The third component corresponds to the influence of the particle that maximizes the FDR. The higher the fitness of the neighbor and the closer its distance to the original particle, the more likely it will influence this particle. A new learning factor is introduced for the FDR component.

In [12], a randomly generated directed graphs are used to define neighborhood where graph links are unidirectional, so a link from *a* to *b* means that *a* considers *b* as a neighbor, but not vice versa. Two methods for modifying the neighborhood structure are tested. The 'random edge migration' method disconnects one side of an edge and connects it to another neighbor, while the 'neighborhood re-structuring' method totally re-initializes the structure after it is kept fixed for a period of time.

In [13], a Hierarchical PSO (H-PSO) version is introduced. In this algorithm, particles are arranged in a hierarchy structure and the best performing particles ascend the tree to influence more particles, replacing relatively worse performing particles which descend the tree. A variant of this algorithm where the structure of the tree itself is made dynamic is presented and tested.

### III. CLUBS-BASED PSO

PSO first models were confined to perceive the swarm as a flock of birds that fly in the search space. The picture of flying birds has limited the imagination of researchers somehow for sometime. Recently, a more broad perception of the swarm as a group of particles, whether birds, humans, or any socializing group of particles began to emerge.

In our proposed C-PSO algorithm, we create 'clubs' for particles analogous to our clubs where we meet and socialize. In our model, every particle can join more than one club, and each club can accommodate any number of particles. Vacant clubs are allowed.

After randomly initializing the particles position and speed in the initialization range, each particle joins a predefined number of clubs, which is known as its 'default membership level', and the choice of these clubs is made random. Then, current values of particles are evaluated and the best local position for each particle is updated accordingly. While updating the particles' speeds, each particle is influenced by its best found position and the best found position by all its neighbors, where its neighborhood is the set of all clubs it is a member of. After speed and position update, the particles' new positions are evaluated and the cycle is repeated.

While searching for the global optima, if a particle shows superior performance compared to other particles in its neighborhood, the spread of the strong influence by this particle is reduced by reducing its membership level and forcing it to leave one club at random to avoid premature convergence of the swarm. On the other hand, if a particle shows poor performance, that it was the worst performing particle in its neighborhood, it joins one more club selected at random to widen its social network and increase the chance of learning from better particles.

The cycle of joining and leaving clubs is repeated every time step, so if a particle continues to show the worst performance in its neighborhood, it will join more clubs one after the other until it reaches the maximum allowed membership level. While the one that continues to show superior performance in every club it is a member of will shrink its membership level and leave clubs one by one till it reaches the minimum allowed membership level.

During this cycle of joining and leaving clubs, particles which no longer show extreme performance in its neighborhood, either by being the best or the worst, go back gradually to default membership level. The speed of going back to default membership level is made slower than that of diverting from it due to extreme performance. The slower speed of regaining default membership level allows the particle to linger, and adds some stability and smoothness to the performance of the algorithm. A check is made every rr (**r**etention **r**ation) iterations to find the particles that have membership levels above or below the default level, and take them back one step towards the default membership level if they do not show extreme performance.

We replace the static inertia weight which controls the momentum of the particle by a uniformly distributed random number in the range (0, *w*). A pseudo code explaining the algorithm is shown below.

```
begin
Initialize particles and clubs
while (termination condition = false)
  do
    evaluate particles fitness: f(x)
    update P
    for (i = 1 to number of particles)
      g_i = best of neighbors_i
      for d = 1 to number of dimensions
        v_id = w×rand_1×v_id + lrn_1×rand_2×(p_id − x_id)
             + lrn_2×rand_3×(g_id −x_id)
        x_id = x_id + v_id
      next d
    next i
    update neighbors
    for j = 1 to number of particles
      if (x_j is best of neighbors_j) and
      (|membership_j| > min_membership)
        leave random club
      end if
      if (x_j is worst of neighbors_j) and
      (|membership_j| < max_membership)
        join random club
      end if
      if (|membership_j| ≠ default_membership)
      and (remainder(iteration/rr) = 0)
        update membership_j
      end if
    next j
  end do
  iteration = iteration + 1
  evaluate termination condition
end while
```

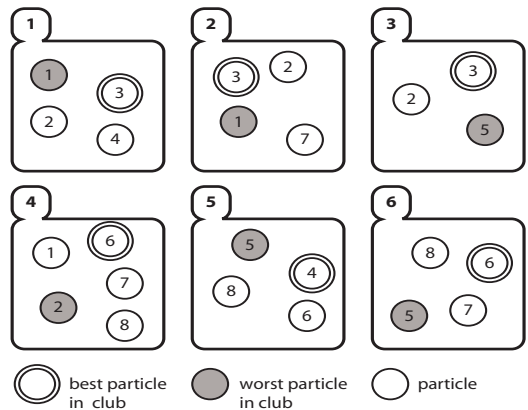

Fig. 1. A snapshot of clubs during a simulation of the C-PSO algorithm

Where P is local best position, neighbors$_i$ is the set of particle i neighbors, membership$_i$, |membership$_i$| are the set of clubs that particle i is a member of and the size of this set respectively. rand$_{1,2,3}$ are three independent uniformly distributed random numbers in the range (0, 1).

Fig. 1 shows a snapshot of the clubs during an execution of the C-PSO algorithm. In this example, the swarm consists of 8 particles, and there are 6 clubs available for them to join. Given the previous pseudo code, and that the minimum, default and maximum membership levels are 2, 3 and 5 respectively, the following changes in membership will happen to particles in Fig. 1 for the next iteration which is a multiple of rr:

1. Particle3 will leave club1, 2 or 3 because it is the best particle in its neighborhood.
2. Particle5 will join club1, 2 or 4 because it is the worst particle in its neighborhood.
3. Particle2 will leave club1, 2, 3 or 4, while particle4 will join club2, 3, 4, or 6 to go one step towards default membership level because they do not show extreme performance in their neighborhood.

*Flow of influence:*

The flow of influence or how the effect of the best performing particles spreads and affects other particles in the swarm is critical to the performance of all PSO algorithms. If the influence spreads quickly through the swarm, they get strongly attracted to the first optimum they find, which is a local optimum in most cases. On the other hand, if the influence spreads slowly, the particles will go wandering in the search space and will converge very slowly to the global or a local optimum.

In order to study the effect of different default membership levels on the flow of influence we do the following experiment. We create a swarm of 20 particles. Clubs membership is assigned randomly but every particle joins exactly *m* of total 100 clubs. The membership level *m* is kept fixed for every single run, so best and worst performing particles do not leave or join clubs. All the particles are initialized to random initial positions in the range [1000 2000]$^n$ except for one particle which is initialized to [0]$^n$. The value of each particle which we want to minimize is simply the sum of its coordinate position values.

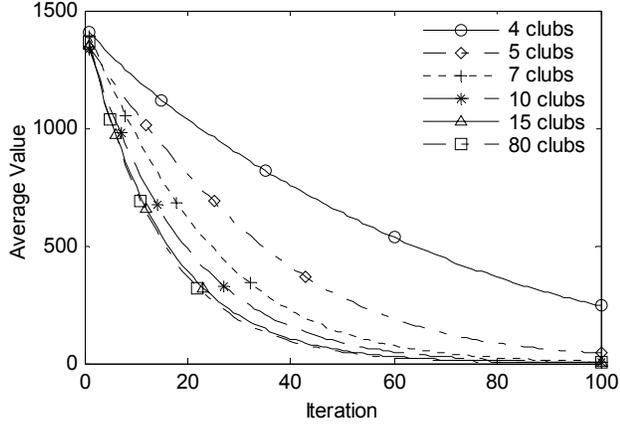

Fig. 2. Effect of different default membership levels on the speed of the flow of influence

The flow of influence for some default membership levels is shown in Fig. 2. The average value of all particles is shown against search progress. The average value of particles decreases because they are influenced by the best performing particle which has a value of '0'. So, rapid decrease of the average value indicates faster flow of influence speed. It is clear that the flow of influence speed monotonically decreases with decreasing default membership levels.

## IV. EXPERIMENTS

The purpose of this paper is to test and analyze the effect of the dynamic social network employed in the proposed C-PSO algorithm on its performance and compare it with the performance of other PSO algorithms which have static social networks. We use five well known benchmark problems presented in TABLE I. The first two functions are simple unimodal functions. They test the ability of the optimizers to deal with smooth landscapes. The next three functions are multimodal functions containing a considerable number of local minima where the algorithm may fall into, so these functions test the ability of the algorithm to escape these traps.

We compare the performance of the different optimizers using two criteria which were used in [13]. The first one is the ability to escape local minima, and is measured by the degree of closeness to the global optimum the optimizer achieves after a long number of iterations. The second one is the convergence speed, which is measured by the required number of iterations to achieve a certain degree of closeness to the global optimum in the evaluation space.

Using these metrics on the five benchmark functions, we compare three versions of the C-PSO with different default membership levels of 10, 15 and 20 from a total number of 100 clubs, along with PSO-g and PSO-l.

The three default membership levels are chosen based on initial empirical results. It was found that lower membership levels decrease the speed of flow of influence, as shown in the previous section, which was reflected on slow conver-gence. While higher membership levels cause premature convergence.

TABLE I
BENCHMARK FUNCTIONS

| Sphere (unimodal) | $f_1(x) = \sum_{i=1}^{n} x_i^2$ |
|---|---|
| Rosenbrock (unimodal) | $f_2(x) = \sum_{i=1}^{n-1} \left[ 100(x_{i+1} - x_i^2)^2 + (x_i - 1)^2 \right]$ |
| Rastrigin (multimodal) | $f_3(x) = \sum_{i=1}^{n} \left[ x_i^2 - 10\cos(2\pi x_i) + 10 \right]$ |
| Schaffer f6 (multimodal) | $f_4(x) = 0.5 + \dfrac{\sin^2\left(\sqrt{x^2 + y^2}\right) - 0.5}{\left(1 + 0.001 \times (x^2 + y^2)\right)^2}$ |
| Achley (multimodal) | $f_5(x) = -20\exp\left(-0.2\sqrt{\dfrac{1}{n}\sum_{i=1}^{n} x_i^2}\right) - \exp\left(\dfrac{1}{n}\sum_{i=1}^{n}\cos(2\pi x_i)\right) + 20 + e$ |

TABLE II
PARAMETERS FOR BENCHMARK FUNCTIONS

| Function | Dim. | Init. range | $V_{max}$ | $w$ C-PSO |
|---|---|---|---|---|
| Sphere | 30 | $[-100; 100]^n$ | 100 | 1.2 |
| Rosenbrock | 30 | $[-30; 30]^n$ | 30 | 1.2 |
| Rastrigin | 30 | $[-5.12; 5.12]^n$ | 5.12 | 1.4 |
| Schaffer's f6 | 2 | $[-100; 100]^n$ | 100 | 1.65 |
| Achley | 30 | $[-32; 32]^n$ | 32 | 1.36 |

For all simulation runs we use $lrn_1 = 1.494$, $lrn_2 = 1.494$, which were used in [13] and suggested by [14]. For PSO-g and PSO-l we use $w = 0.729$ as in [13] and [14], while the value of $w$ for C-PSO which reflects the range of the random inertia weight is presented in TABLE II for each problem.

The minimum and maximum allowed membership levels we use are 5 and 33 respectively, while rr = 2. We use a swarm of 20 particles for all simulation runs.

The particles position and speed are randomly initialized in the ranges shown in TABLE II depending on the benchmark problem used. The absolute speed values for particles are kept within the $V_{max}$ limit for all dimensions during simulation. On the other hand, particles movements are not restricted by boundaries, so particles may go beyond the initialization range and take any value.

Every simulation run was allowed to go for 10000 iterations, and each simulation has been repeated 50 times. All simulation runs were executed using MATLAB® R2006a.

## V. RESULTS

Each graph presented in this section represents the average of the 50 independent simulation runs for all optimizers unless otherwise stated.

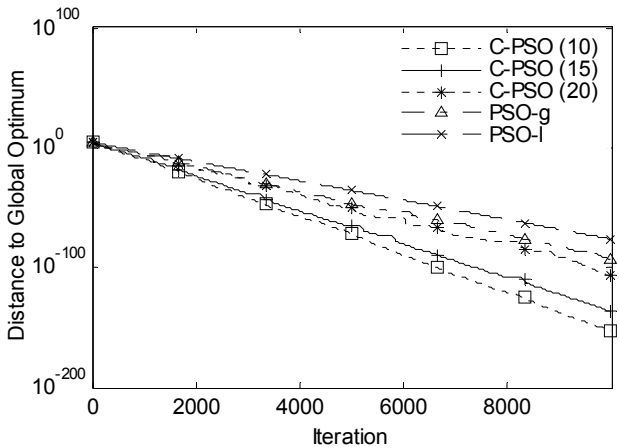

Fig. 3. Sphere—Closeness to global optimum for
C-PSO (10, 15, 20), PSO-g and PSO-l

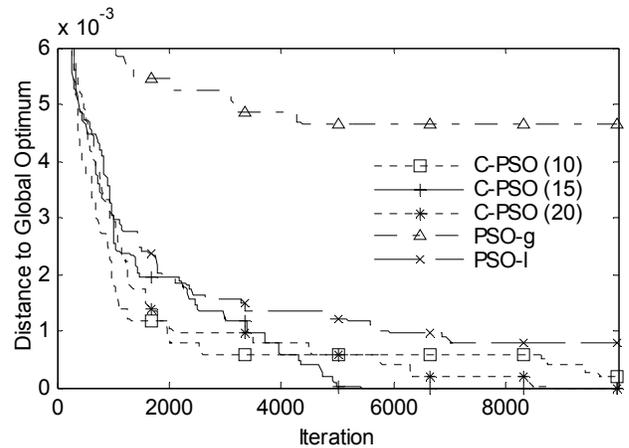

Fig. 6. Schaffer's f6—Closeness to global optimum for
C-PSO (10, 15, 20), PSO-g and PSO-l

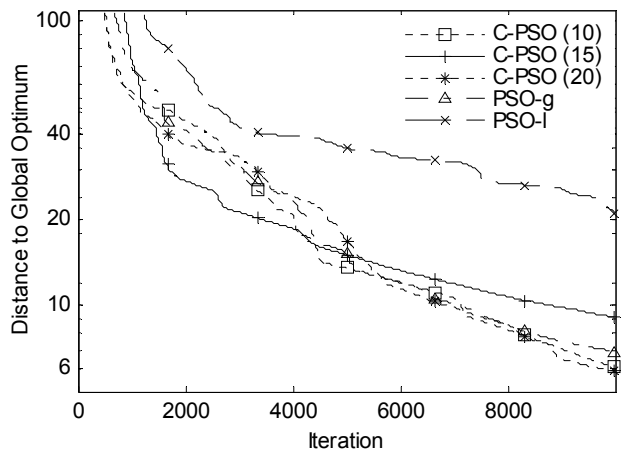

Fig. 4. Rosenbrock—Closeness to global optimum for
C-PSO (10, 15, 20), PSO-g and PSO-l

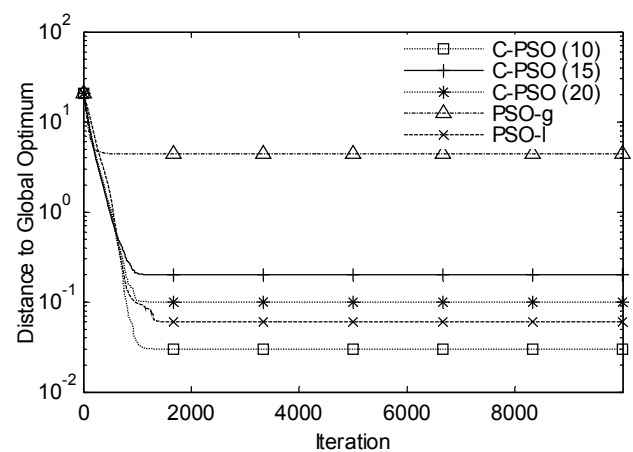

Fig. 7. Ackley—Closeness to global optimum for
C-PSO (10, 15, 20), PSO-g and PSO-l

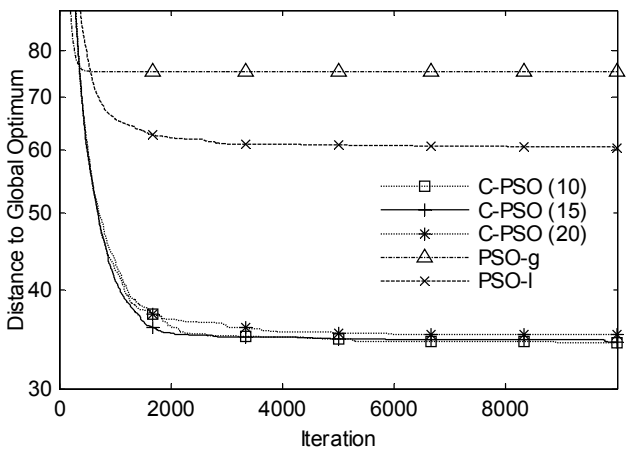

Fig. 5. Rastrigin—Closeness to global optimum for
C-PSO (10, 15, 20), PSO-g and PSO-l

*A. Escaping Local Minima:*

We begin with the first criterion which is the ability of the algorithm to escape local minima. The Sphere and Rosenbrock problems have the lowest number of local minima. Their unique minimum makes them the easiest of the five benchmark problems in finding the global minimum.

For the Sphere problem as shown in Fig. 3, all C-PSO versions managed to finish closer to the unique minimum than PSO-g and PSO-l, and the lower the membership level the faster the algorithm converges. The PSO-l algorithm was the worst performer followed by PSO-g.

For the Rosenbrock problem presented in Fig. 4, C-PSO (10, 20) and PSO-g show very close performance, though PSO-g is little behind them. C-PSO (15) follows them by a short distance, while PSO-l is the worst of all, lagging behind by a relatively long distance.

As shown in Fig. 5 and Fig. 6, we can see that all C-PSO versions perform better than both PSO-g and PSO-l for the Rastrigin and Schaffer's f6 test problems respectively. In both of them, PSO-g gives the worst performance and converges prematurely in the Rastrigin problem, followed by PSO-l as the second worst. All C-PSO versions give similar performance for the Rastrigin problem as we can hardly distinguish them.

As shown in Fig. 7 for the Ackley problem. C-PSO (10) outperforms all the other algorithms followed by PSO-l, C-PSO (20, 15) respectively, while PSO-g suffers premature convergence again and falls by a long distance behind. The distances to global optima after 10000 iterations of the optimizers are shown in TABLE III.

TABLE III
DISTANCES TO GLOBAL OPTIMA AFTER 10000 ITERATIONS

| Algorithm | Sphere | Rosen. | Rastrigin | Schaffer | Ackley |
|---|---|---|---|---|---|
| PSO-l | 7.4e-77 | 21.08 | 60.37 | 0.00077 | 0.06 |
| PSO-g | 4.2e-93 | 6.88 | 75.32 | 0.00466 | 4.45 |
| C-PSO (20) | 1.3e-107 | **5.92** | 35.10 | **0** | 0.10 |
| C-PSO (15) | 5.4e-137 | 9.11 | 34.34 | **0** | 0.20 |
| C-PSO (10) | **1.1e-152** | 6.07 | **34.30** | 0.00019 | **0.03** |

As expected, the performance of PSO-g and PSO-l depend on the problem they optimize. For the first two problems which have a single optimum, PSO-g performs better than PSO-l, as all the particles get strongly attracted to the unique optimum due to the fully connected social network in PSO-g. On the other hand, PSO-l goes wandering and converges slowly.

For the last three problems, which have many local optima, PSO-l outperforms PSO-g. The partially connected social network of PSO-l creates many points of attraction for the particles in the swarm that help them escape some local optima compared to PSO-g.

Unlike PSO-g and PSO-l, C-PSO performance is much less problem dependent. C-PSO (10) outperformed both PSO-g and PSO-l for all problems. The results obtained for the Sphere problem were unexpected. The unique minimum and the non-deceptive landscape of the problem make a perfect match with PSO-g. The fully connected social network should do a better job in attracting the particles to the unique global minimum than any other social network.

These results necessitated further investigation into the behavior of the optimizers and specially the flow of influence through the swarm in unimodal and multimodal problems. So we do the following experiment.

*B. Further Investigation of Optimizers' Behaviors:*

We run three optimizers, C-PSO (10), PSO-l and PSO-g on the unimodal Rosenbrock test problem and the multimodal Rastrigin problem. During the simulation run we record the index of the best performing particle in the swarm for each iteration. We use the same parameters used previously for the first criterion.

Fig. 8 and Fig. 9 show the best performing particles in C-PSO (10) (top), PSO-l and PSO-g (bottom) for the Rosenbrock and Rastrigin problems respectively. For each plot, the index of particles (20 particles) is drawn against the number of iterations elapsed. A dot at (13, 5000) indicates that 'particle 13' has the global best value in the swarm during 'iteration number 5000'.

First, considering Rosenbrock problem shown in Fig. 8. The status of being the best performing particle in the case of C-PSO is almost uniformly distributed over all particles, once a particle finds a good solution, another particle finds a better one.

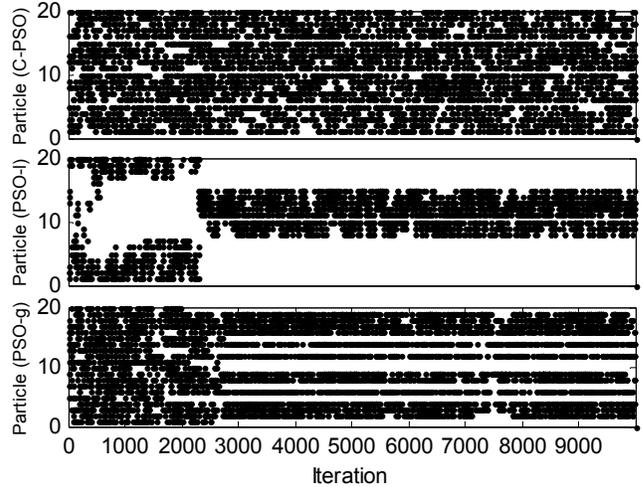

Fig. 8. Best particle in the swarm for Rosenbrock problem using C-PSO(10) (top), PSO-l, and PSO-g (bottom)

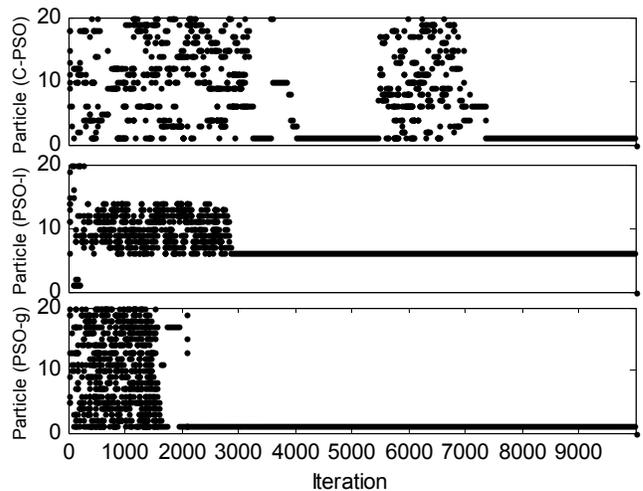

Fig. 9. Best particle in the swarm for Rastrigin problem using C-PSO(10) (top), PSO-l, and PSO-g (bottom)

A reason for this behavior is that once a particle finds the good solution it becomes the best particle in the swarm, making it the best in its neighborhood as well. The particle shrinks its membership level one by one and reduces its influence on other particles accordingly. Neighbors of this superior particle will carry its influence to other clubs they are member of, so other particles are still indirectly guided by it, but are more free to find a steeper way down the hill to the global optimum. Once a particle finds it, it becomes the new best particle and continues or starts shrinking its membership level (because it may become the best in its neighborhood before becoming the global best). The particles which are no longer the best in their neighborhood regain their default membership level to increase their chance in learning from better particles to become the new global best, and the cycle continues.

On the other hand, the best particle status in PSO-l goes bouncing between two particles on the ring (note that particle 1 is connected to particle 20) as shown in Fig. 8 (middle). The particles outside this arc search totally inefficient regions of the search space. This is clear from the fact that none of them shows up even once as the best particle in the swarm for the last 7500 iterations. This clustering mechanism may help the algorithm to overcome local optima in multimodal problems, but in unimodal problems it has detrimental effect.

The reason that PSO-g algorithm came second to C-PSO in both unimodal problems is clear in Fig. 8 (bottom). After around 2600 iterations, 12 particles acted as guides for the other 8 particles and literally dragged them behind. None of the 8 particles showed superior performance till the end of the 10000 iterations.

Second, we consider Rastrigin problem presented in Fig. 9. This multimodal problem requires diversity in the swarm and a clever social network to overcome local optima. The property of fully connected social network in PSO-g provokes all the particles to jump to the best found position by all particles in the swarm.

This makes the first 1500 iterations for PSO-g look almost the same for both unimodal and multimodal problems. But after these 1500 iterations the algorithm prematurely converges in the case of the multimodal Rastrigin problem as shown in Fig. 9 (bottom).

On the other hand, PSO-l algorithm presented in Fig. 9 (middle) maintains its diversity for a longer period than PSO-g does. Along with its clustering property explained earlier, it manages to escape local optima to some extent and get closer to global optimum than PSO-g can get.

Finally for the C-PSO optimizer as shown in Fig. 9 (top), the algorithm maintains diversity longer than PSO-g and PSO-l do. Moreover, the best performing particle status is distributed over the particles, unlike PSO-l, and the particles do not jump over the best particle once it emerges. This can be seen as the particles create more clusters than in the case of PSO-l and PSO-g. These clusters represent local optima found by the particles. The most interesting result found is the ability of the C-PSO to explore new regions after a period of stagnation. We can see how C-PSO finds better regions at around iteration 5200 after it has stagnated for nearly 2000 iterations.

An explanation for this behavior is that the best performing particles in their neighborhood create different points of attraction for the particles. The particles are grouped according to their clubs' membership and search the space around these points of attraction. At the same time, the worst particles on their neighborhood expand their membership and bridge the influence between different groups of searching particles. If a searching group finds a better solution, its influence is transmitted over the bridge acting particles to other groups and diverts them from searching inefficient regions indefinitely. They start searching for other optima which could be better than the best one found and create different points of attraction, and the cycle goes on.

*C. Convergence Speed:*

The second criterion to be considered is the convergence speed of the algorithms. As explained earlier, it is being measured by the number of iterations the algorithm takes to reach a certain degree of closeness to the global optimum.

TABLE IV. NUMBER OF ITERATIONS NEEDED TO REACH A CERTAIN DEGREE OF CLOSENESS TO GLOBAL OPTIMUM FOR THE FIVE OPTIMIZERS. (BEST VALUES ARE BOLD FACED)

| Algorithm | Avg. | Med. | Max. | Min. | Suc.% |
|---|---|---|---|---|---|
| Sphere – (Closeness = 0.0001) | | | | | |
| PSO-l | 1030.8 | 1036 | 1103 | 965 | **100** |
| PSO-g | 684.88 | 672 | 1012 | 489 | **100** |
| C-PSO (20) | 611.68 | 571 | 1057 | 421 | **100** |
| C-PSO (15) | 528.18 | **506.5** | 711 | **417** | **100** |
| C-PSO (10) | **518.14** | 513.5 | **652** | 443 | **100** |
| Rosenbrock – (Closeness = 100) | | | | | |
| PSO-l | 1429.6 | 907 | 7465 | 604 | 98 |
| PSO-g | 874.3 | 425 | 6749 | 251 | **100** |
| C-PSO (20) | 697.3 | 424 | 4537 | 240 | **100** |
| C-PSO (15) | **569** | 473 | **1605** | 218 | 98 |
| C-PSO (10) | 725.8 | **376** | 6016 | 226 | 98 |
| Rastrigin – (Closeness = 50) | | | | | |
| PSO-l | 1695.7 | 1068 | 8015 | 500 | 26 |
| PSO-g[a] | 250 | 221 | 313 | 216 | 6 |
| C-PSO (20) | 813.9 | 702 | 3396 | **254** | 88 |
| C-PSO (15) | **695.4** | **597.5** | 1829 | 262 | 88 |
| C-PSO (10) | 753.3 | 667 | 1932 | 299 | **96** |
| Schaffer's f6 – (Closeness = 0.001) | | | | | |
| PSO-l | 1076.2 | 422 | 7021 | 84 | 92 |
| PSO-g | **791.1** | **279.5** | **4276** | 60 | 52 |
| C-PSO (20) | 1138.3 | 524 | 8462 | 80 | **100** |
| C-PSO (15) | 1120.1 | 432 | 4966 | 88 | **100** |
| C-PSO (10) | 945.6 | 401 | 9668 | **48** | 98 |
| Ackley – (Closeness = 0.01) | | | | | |
| PSO-l | 968.2 | 954.5 | 1531 | 827 | 96 |
| PSO-g[a] | 499 | 499 | 499 | 499 | 2 |
| C-PSO (20) | 831.1 | 806 | 1148 | 610 | 92 |
| C-PSO (15) | **800.6** | **793.5** | **1141** | **570** | 84 |
| C-PSO (10) | 863.7 | 841 | 1151 | 672 | **98** |

[a] Not considered in comparison due to its very low success rate

This number of iterations should be small enough to reflect the ability of the algorithm to converge rapidly, and not its ability to escape local optima and achieve better values at later stages of the run. On the other hand, the closeness value chosen should lie close enough to the global optimum to be efficient in practical applications. We choose closeness values for the five benchmark problems that are satisfied by most algorithms around the range of [500, 1000] iterations. These closeness values are shown in TABLE IV next to problem names.

The figures presented in TABLE IV are compiled from the same results data set collected for the first criterion. They represent the Average, Median, Maximum, Minimum and Success rate of 50 independent simulation runs for the five optimizers. Only data of successful runs were used to evaluate these values, so the sample number is not the same for all figures.

We can see from TABLE IV that C-PSO (15) achieves the overall best results. For the Sphere problem, all the algorithms achieve the desired closeness in every single run, though C-PSO (10, 15) come ahead of them. The situation is similar in the second unimodal Rosenbrock problem, however the success rate is lower for PSO-l and C-PSO (10, 15).

Moving to multimodal problems, we notice how PSO-g shows poor performance in reaching the closeness values. For Rastrigin and Ackley problems, it only succeeds in six and two percent of the runs respectively, compared to much higher success rates in all C-PSO versions.

C-PSO (15) outperform all the other algorithms for Rastrigin and Ackley problems, except for the Rastrigin problem where it comes second to C-PSO (20) regarding the minimum number of iterations in the 50 samples. We do not consider PSO-g in our comparison for Rastrigin and Ackley problems due to its very low success rate.

Finally for Shaffer's f6 problem, PSO-g achieved the best results for the mean, median and maximum number of iterations. It should be noted however that it has a low success rate of 52% which is almost half the success rate for all C-PSO versions. This low success rate makes it unreliable in practical applications.

## VI. CONCLUSION

Particle swarm optimizers are very sensitive to the shape of their social network. Both PSO-g and PSO-l lack the ability of adapting their social network to the landscape of the problem they optimize.

The proposed C-PSO algorithm overcomes this problem. The dynamic social network of the optimizer shrinks the membership level of superior particles to reduce their influence on other particles, while expanding the membership level for the worst particles to increase their chance in learning from better particles.

C-PSO versions achieved better results than PSO-l and PSO-g either in escaping local optima or in convergence speed to global optima for almost all benchmark problems we considered.

Further investigations have shown that the dynamic social network allowed particles to be guided indirectly by the superior particles, while searching for better solutions more freely than the case of PSO-g. It was shown using empirical results that C-PSO is able to explore and find better regions in the search space during periods of stagnation, making it attractive for use in multimodal problems.